# You want to survive the data deluge? Be careful: Computational Intelligence will not be your rescue boat

Emanuel Diamant
VIDIA-mant
Kiriat Ono 5510801, Israel
emanl.245@gmail.com

*Abstract*— We are at the dawn of a new era, where advances in computer power, broadband communication and digital sensor technologies have led to an unprecedented flood of data inundating our surrounding. It is generally believed that means such as Computational Intelligence will help to survive these tough times. However, these hopes are improperly high. Computational Intelligence is a surprising composition of two mutually exclusive and contradicting constituents that could be coupled only if you disregard and neglect their controversies: "Computational" implies reliance on data processing and "Intelligence" implies reliance on information processing. Only those who are indifferent to data-information discrepancy can believe that such a combination can be viable. We do not believe in miracles, so we will try to share with you our reservations.

## I. INTRODUCTION

Recent advances in sensor technologies, explosive growth of computing power, proliferation of broadband internet communication, development of innovative content sharing services over the World Wide Web have led to an unprecedented flood of data inundating our surrounding. We are at the dawn of a new era, the era of Big data, where huge amounts of raw data come down upon us with a mighty power of data deluge.

It goes without further saying that in such circumstances traditional practice of human-centered management and handling of such data volumes does not hold anymore and has to be entrusted to a machine (a computer as we usually call it today). However, it is self-evident that such a computer has to possess many human-like cognitive abilities, which underpin understanding, analysis, and interpretation of the incoming data streams. That is the reason and the purpose of the Computational Intelligence (CI) advent.

The clarity of exposing the roots of CI constituents does not remove the uncertainty inherited from the parents' family traits. "Computational" comes from the early days of the past century mid, when computer has become an indispensable part of our life and all around has become computational: Computational neuroscience, Computational genomics, Computational chemistry, Computational ecology, Computational linguistics, Computational intelligence, and so on.

The "Intelligence" was inherited from the Artificial Intelligence (AI) branch of science, a fundamental concept that was invented at about the same time (the time of computer dawn) at the Dartmouth College meeting in the summer of 1956. Four brilliant scientists (J. McCarthy, M.L. Minsky, N. Rochester, and C.E. Shannon) have worked out and put into operation an excellent idea. But, despite their prominence, the founding fathers failed to assess the complexity of the task. The term "Intelligence" was left undefined. It was assumed that the best-known manifestation of intelligence is human intelligence; therefore, AI's aim was affirmed as human intelligence replication. It was also assumed that, because the brain is the core of intelligence and the brain is busy with information processing, intelligence should be defined as a product of information processing. Hence, one of the AI's destinations has become information processing. And that is what the CI has inherited from the AI family treasures.

On the other hand, the computational roots of CI (as it was explained just above) imply that CI's destination is data processing, because computer, by definition, is a data processing device.

This contradiction (data processing versus information processing) has never bothered the CI designers. The blurred delineation between the terms data and information, the ubiquitous interchangeable use of them was inspired by the Shannon's "Mathematical Theory of Communication", [1], and the Information Theory embedded in it. For a long time, during all the second half of the past century, Shannon's Information Theory was the dominant research paradigm of the scientific community. The original aim of the theory was to solve a purely technical problem: to increase the performance of a communication system. In his theory, Shannon defines information in terms of signal's statistical properties and the uncertainty of receiving a particular signal among those that are possible. The theory has explicitly linked information notion with data and set aside any discussion about signal's value or meaning.

However, in today's modern sciences, meaning and semantic aspects of a message have become of a paramount importance, and the disregarded distinction between data and information cannot be tolerated anymore. Therefore, it will be our duty to clarify the matters before we proceed with further analysis of CI's ability to serve us in the circumstances of Big data deluge.

## II. DANA AND INFORMATION: WHAT IS THE DIFFERENCE

As it was said, Shannon defines information as the entropy of a discrete set of probabilities, as an opportunity to reduce uncertainty of a received data transmit. My approach to information relies on the Kolmogorov's view on the subject [2].

A slightly modified and an extended version of Kolmogorov's description sounds today (in my words) like this: **"Information is a linguistic description of structures observable in a given data set"**.

To make the scrutiny into this definition more palpable I propose a digital image to be considered as a given data set.

A digital image is a two-dimensional set of data elements called picture elements or pixels. In an image, pixels are distributed not randomly, but, due to the similarity in their physical properties, they are naturally grouped into some clusters or clumps. I propose to call these clusters **primary or physical data structures**.

In the eyes of an external observer, the primary data structures are further arranged into more larger and complex agglomerations, which I propose to call **secondary data structures**. These secondary structures reflect human observer's aptitude to the grouping of primary data structures, and therefore they could be called **meaningful or semantic data structures**. While formation of primary (physical) data structures is guided by objective (natural, physical) properties of the data, the ensuing formation of secondary (semantic) data structures is a subjective process guided by human conventions and habits.

As it was said, **Description of structures observable in a data set should be called "Information".** In this regard, two types of information must be distinguished – **Physical Information and Semantic Information**. They are both language-based descriptions; however, physical information can be described with a variety of languages (recall that mathematics is also a language), while semantic information can be described only by means of natural human language. (More details on the subject you can find in [3]).

Those, who will go and look in [3], would discover that every information description is a top-down evolving coarse-to-fine hierarchy of descriptions representing various levels of description complexity (various levels of description details). Physical information hierarchy is located at the lowest level of the semantic hierarchy. The process of sensor data interpretation is reified as a process of physical information extraction from the input data, followed by an attempt to associate this physical information (about the input data) with physical information already retained at the lowest level of a semantic hierarchy. If such an association is achieved, the input physical information becomes related (via the physical information retained in the system) with a relevant linguistic term, with a word that places the physical information in the context of a phrase, which provides the semantic interpretation of it. In such a way, the input physical information becomes named with an appropriate linguistic label and framed into a suitable linguistic phrase (and further – in a story, a tale, a narrative), which provides the desired meaning for the input physical information.

The segregation between physical and semantic information is the most essential insight about the nature of information. Another insight is that, because of the subjective nature of semantic information, its creation cannot be formalized. Semantic information hierarchy, thus, cannot be learned and has to be provided to the system always from the outside, always as a gift, a grant, an offering. The next important outcome from the definition given above is the understanding that information descriptions are always reified as a string of words, a piece of text, a narrative.

Bearing in mind all these new peculiarities, we can proceed to further revision of information processing implications for the research into the Big data challenges.

## III. SORRY TO INFORM YOU

It is generally acknowledged that in the Big data analytics CI is the most promising technique used to face complex modeling, prediction, and recognition tasks. There is an ample literature devoted to the Big data analysis problems. References [4], [5], [6], [7] are given only to reflect the ubiquity of this reality.

As follows from this literature, Big data analysis revolves around a standard package of procedures – to discover patterns in data, to select characteristic features of these patterns to facilitate the pattern recognition processes, to reveal relations among the patterns that will be used for further data interpretation, understanding, and decision making. Knowledge discovery and information retrieval (from data) are also frequently mentioned as prime big data analysis goals.

Vigilant readers will easily distinguish the striking similarity between the steps of our information descriptions building process and the treads of big data intelligent processing. Both commence with data clustering (detection, recovery, revealing) procedure. In CI that is called data patterns discovery, in our approach, it is called data structures delineation. In CI, they call this data structure information; in our approach, we call it "the primary data structure outlining", which is used for a further structure description creation. Only this structure description can be called information, more precisely – physical information. And for this physical information a suitable semantic interpretation will be subsequently retrieved from the system's knowledge base (and that will be the desired semantic information used for decision making and behavior forecasting).

CI tools developers know nothing about the duality of information, about the subjective nature of the semantic information (which is a convention, a mutual agreement among the observers, and therefore, it is an observers property and not a property of the data). They do not recognize the presence of secondary structures and the subjective rules of their formation. Therefore, CI developers, after delineating first elementary patterns, use them as building blocks for creation more complex patterns, called "features", which then are further arranged in even more complex patterns, called "objects", for which they are trying to derive and to formulate an interpretation. That does not work. Semantic information cannot be derived from the physical information! So all their hard attempts are in vein.

An example and an evidence that data, by itself, (the characteristic data features) do not take part in semantic information processing (in data patterns recognition) can be drawn from the following facts that are well known to all of you from your own experience:

- We get the meaning of a written word irrelevant to the font size of the letters or their style (irrelevant to basic data features).
- We recognize equally well a portrait of a known person on a huge-size advertising billboard, on a magazine front page, or on a postage stamp – perceptual pictorial information (physical information) is dimensionless, (while data features are not).
- We get the meaning of a scene irrelevant to its illumination. We look on the old black-and-white photos and we do not perceive the lack of colors (a prime image data feature).
- The same is true for voice perception and spoken utterance understanding – we understand what is being said irrelevantly to who is speaking (a man, women, and a child). Irrelevant to the volume levels of the speech (loudly or as a whisper).
- Blind people read Brail-style writings irrelevant to the size or the form of the touched Brail code, irrelevant to their perceptible temperature.

Again – physical information is a **description of structures** observable in a given data set, not the data in those structures. Only "description of structures" take part in further information processing. Original data features are become dissolved in the descriptions and do not take part in a data set (a scene, e.g.) understanding/interpretation process.

To summarize this paragraph it must be said that machine learning, fuzzy logic, evolutionary and genetic algorithms, which CI proudly embraces as its work horses, are all busy with raw data processing (data as their natural input), that is, busy with physical information processing only. And any semantic information cannot be revealed in course of such data processing.

The subjective rules of knowledge base construction put in doubt the claims of the effectiveness of various forms of machine learning approaches implemented for big data analysis (tasks). As it is explained in [3], semantic information hierarchy, which plays the role of a referential knowledge base for input (physical information) interpretation, cannot be gained autonomously. It has to be delivered to the system's disposal from the outside, be granted or given as a gift from an external supplier. It can be modified then and adapted to the system's needs, but it cannot be developed and learned in an autonomous way.

CI design approaches and CI tools development practice are repetitively violating this important rule.

Some remarks must be made considering Knowledge-Information relationships. The commonly accepted view is that these relationships are represented by a hierarchical structure known as DIKW (Data, Information, Knowledge, Wisdom) pyramid [8]. That is, information is always defined in terms of data, knowledge in terms of information, and wisdom in terms of knowledge. In the light of the definition of semantic information (already given in this paper), we can proclaim that the DIKW pyramid is a wrong representation of data – information – knowledge interrelations. That semantic information hierarchy is exactly what is being called the system's knowledge base (hierarchy). Therefore, we can say that knowledge is a memorized semantic information. That is, knowledge is the semantic information retained in the system's memory for the purposes of physical information (contained in the input data) identification and interpretation. The crucially important issue of Knowledge-Information interrelationships is totally overlooked in the contemporary CI design practice.

IV. CONCLUSIONS

Big data storming streams are raging around. To survive the flood, people are preparing their rescue means. But crafting Computational Intelligence as a solution – is a bad idea. Its construction presumes integration of two mutually exclusive and conflicting components – "Computational", which implies reliance on data processing, and "Intelligence", which implies reliance on information processing. People usually do not pay attention to this disparity. Therefore, they take seriously the CI promises, in spite of hidden flaws and faults of CI design philosophy. Another example of such type of mindset disorder is the "Cognitive Computing" innovation. In it, only the order of the constituting parts is different, but in essence they are the same contradicting entities – Cognitive implies information processing and Computing implies data processing. There is no need to remind you that their duties are incompatible.

Rescue vessels of this type will not be of use for you. I do not think that my explanations will be adopted by the brave sailors riding the Big data depths. I am thinking about the treasures that will be lost forever at the sea bottom. But that is the way the world works, isn't it?